\documentclass[runningheads]{llncs}
\usepackage{graphicx}
\usepackage{amsmath,amssymb} 
\usepackage{color}
\usepackage[width=122mm,left=12mm,paperwidth=146mm,height=193mm,top=12mm,paperheight=217mm]{geometry}
\usepackage{wrapfig}
\usepackage{xspace}
\usepackage[inline]{enumitem}
\usepackage[title]{appendix}
\newcommand{\taskAcro}{WIQ\xspace}
\newcommand{\dataAcro}{TIWIQ\xspace}

\makeatletter
\newcommand{\printfnsymbol}[1]{%
  \textsuperscript{\@fnsymbol{#1}}%
}
\makeatother

\begin{document}
\pagestyle{headings}
\mainmatter
\def\ECCV18SubNumber{}  

\title{Answering Visual {\em What-If} Questions:\\From Actions to Predicted Scene Descriptions}

\titlerunning{Answering Visual {\em What-If} Questions}

\authorrunning{M. Wagner et al.}
\author{Misha Wagner\inst{1}\thanks{These authors contributed equally to this work.} \and Hector Basevi\inst{1}\printfnsymbol{1} \and 	Rakshith Shetty\inst{2} \and Wenbin Li\inst{2} \and \newline Mateusz Malinowski\inst{2} \and Mario Fritz\inst{3} \and Ale\v{s} Leonardis\inst{1}}
\institute{University of Birmingham \and Max Planck Institute for Informatics, Saarland Informatics Campus \and CISPA Helmholtz Center i.G., Saarland Informatics Campus}

\maketitle

\begin{abstract}
In-depth scene descriptions and question answering tasks have greatly increased the scope of today's definition of scene understanding. While such tasks are in principle open ended, current formulations primarily focus on describing only the current state of the scenes under consideration. In contrast, in this paper, we focus on the future states of the scenes which are also conditioned on actions. We posit this as a question answering task, where an answer has to be given about a future scene state, given observations of the current scene, and a question that includes a hypothetical action. 
Our solution is a hybrid model which integrates a physics engine into a question answering architecture in order to anticipate future scene states resulting from object-object interactions caused by an action. We demonstrate first results on this challenging new problem and compare to baselines, where we outperform fully data-driven end-to-end learning approaches.
\end{abstract}

\section{Introduction}
While traditional scene understanding involves deriving bottom-up scene representations such as object bounding boxes and segmentation, in recent years alternative approaches such as \emph{scene captioning} and \emph{question answering} have become increasingly popular. These do not strive for a particular type of representation of the input scene, but rather formulate an alternative task that requires a more holistic scene understanding. Such approaches have been very successful and have shown great advances in extracting the semantic scene content by deriving captions and answers about diverse scene elements.

Beyond the estimation of the ``status quo'' of a visual scene, recent deep learning approaches have shown improved capabilities of forecasting scenes into the future. This is particularly useful for autonomous agents (e.g., robots or driving assistants) that have to plan ahead and act safely in dynamically changing environments. 
Recent approaches show extrapolation of complete videos \cite{mathieu2015deep}, edge information \cite{bhattacharyya2016long} or object trajectories \cite{lerer2016learning,mottaghi2016happens}. However, with increasing time horizons and complexity of the scenes, such quantitative predictions become increasingly difficult. In addition, extrapolation of complete image data might be wasteful and overly difficult to achieve.

Furthermore, current work on anticipation and forecasting is typically not interactive, meaning that the agent is acting purely as a passive observer. However, in many real-world applications, an agent is faced with the task of evaluating multiple different potential actions that will cause diverse outcomes. The future is therefore often conditioned on the actions of the agent which is not handled by the state-of-the-art methods.

Therefore, we argue for a qualitative prediction of the future conditioned on an action. We phrase this as the {\it Answering Visual What-If Questions} task, where the answer is conditioned on an observation and question including a hypothetical action. 
This formulation allows us to evaluate a model's forecasting abilities conditioned on a hypothetical action, and at the same time allows for sparse representation of the future where not all details of the scene or object interactions have to be fully modeled.

We provide the first investigation of this challenging problem in a table top scenario where a set of objects is placed in a challenging configuration and different actions can be taken. Dependent on the action, the objects will interact according to the physics of the scene and will cause a certain outcome in terms of object trajectories. The task is to describe the outcome with respect to the action. In order to address this problem we couple the question answering approach with a physics engine -- resulting in a hybrid model. While several parts of the method, such as inferring the action from the question and predicting the answer, are data-driven, the core aspect of our model that predicts future outcomes is model-based (using a physics engine). 

\noindent
\paragraph{\bf The contributions of this paper are as follows:}
\begin{itemize}
\item We define a new challenge called Visual \emph{What-If} Question answering~(WIQ)
that brings together question answering with anticipation.
\item We propose the first dataset for the \taskAcro  task based on table-top interactions between multiple objects called \dataAcro.
\item We propose the first hybrid model that uses a physics engine together with a question answering architecture.
\end{itemize}

\section{Related Work}
\textbf{Learning Physics and Future Predictions:}
Coping with the physical world by predicting how  objects interact with each other using rules of physics is among the pillars of human intelligence.
This type of intuitive understanding of physics, often referred to as ``intuitive physics'' \cite{mccloskey1983intuitive}, is also becoming of interest to machine learning researchers. The ``NeuroAnimator'' is among the first learning-based architectures trained to simulate physical dynamics based on observations of physics-based models \cite{grzeszczuk1998neuroanimator}. Although the ``NeuroAnimator'' is mainly motivated  by efficiency, 
others have realized that learning-based architectures may be key components to learn the whole spectrum of physical reasoning that humans possess. For instance, \cite{battaglia2013simulation} argues that a cognitive mechanism responsible for physical reasoning may resemble an engine that simulates complex interactions between different physical objects, and can be implemented by ``graph neural networks'' \cite{battaglia2016interaction,watters2017visual} or by an engineered physics engine~\cite{wu2015galileo}. In this work, our hybrid model also uses a physics engine, but unlike~\cite{wu2015galileo} we are less interested in inferring latent physics properties of two objects from videos, but rather in a forward model of physics for the purpose of answering questions. 
A complementary line of research has shown that convolutional neural networks (CNN) are capable to some extent of physical reasoning such as stability prediction \cite{lerer2016learning,li2017visual,li2016fall}, or future frame synthesis from RGB-D input \cite{ranzato2014video,mathieu2015deep,bhattacharyya2016long} or even static images \cite{mottaghi2016newtonian,mottaghi2016happens}.
These approaches to 
physical intelligence
focus on testing this understanding either by trying to extrapolate sensory data into the future (predicting video frames) or by inferring individual properties (predicting stability or physics properties). In contrast, we propose to achieve qualitative physical understanding, where we want the model to have general understanding of physical processes but not necessarily the ability to make precise prediction.
\newline
\textbf{Visual Question Answering (Visual QA):} 
This is a recently introduced research area \cite{malinowski2014multi,malinowski2014towards} that attempts to build and understand if machines can learn to explain the surrounding environment only based on questions and answers. Since then, the community has seen a proliferation of various datasets \cite{geman2015visual,ren2015imageqa,yu2015visual,zhu2016visual7w,tapaswi2016movieqa,agrawal2017don,kafle2018dvqa}, including the most popular VQA \cite{antol2015vqa}, as well as numerous methods \cite{malinowski2014multi,ren2015imageqa,malinowski2017ask,yang2016stacked,fukui2016multimodal,hu2017learning,santoro2017simple}. 
Although most of the questions involve static scenes, and are either related to objects, attributes, or activities, there are some that require understanding of physics at
the ``intuitive level''. Consider even such seemingly simple question as ``What is on the table?''. To interpret this question, understanding of ``on'' is needed, and this involves physical forces such as gravity. In spite of the existence of such questions in the aforementioned datasets, due to lack of interactions, it is hardly possible the learnt models can really understand them, and likely they only rely on visual correlations. In our work, through the interactions, and exploitation of physics, we can train architectures that, we hypothesize, can model physical interactions between objects.
\newline
\textbf{Simulations and Machine Learning:}
Since it is difficult to generate realistic data that includes complex physical interactions, most approaches either rely on short videos with limited exposition to physics \cite{ranzato2014video,mathieu2015deep,bhattacharyya2016long} or on synthetically generated data \cite{battaglia2016interaction,watters2017visual,li2016fall,ehrhardt2017taking}. This problem of lacking good realistic environments with rich physical interactions also governs the research on reinforcement learning \cite{sutton1998reinforcement}, where the community often relies on game-like environments \cite{beattie2016deepmind,kempka2016vizdoom,airsim2017fsr,wu2018building}. Since there is no publicly available realistic environment that has rich enough physical interactions that we are interested in, we build a dataset consisting of 3D scenes, with physical interactions, and with realistically textured objects.

\section{Visual \emph{What-If} Questions (WIQ) Task}
While Visual QA evaluates the scene understanding of a passive agent, this is not
sufficient for an active agent 
that needs to anticipate the consequences of its
actions and communicate about them.
To study this aspect of scene understanding, we propose the task of answering ``what-if'' questions
pertaining to a visual scene.
The agent is shown an input scene with multiple objects and is given a hypothetical action
description.
It then has to describe what happens to different objects in the scene, given that it
performs the specified action.
This amounts to answering questions of the form ``If I perform action \emph{A}, what happens to
object \emph{X}?''.
To answer such questions the agent has to
parse the natural language description of the action to infer the action
type and target object on which the action is applied, along with the corresponding parameters such as the initial force.
Then the agent needs to anticipate the consequences of this action on different objects in the
scene, and finally verbalize its predictions.
This combines the challenges involved in the standard VQA task~\cite{antol2015vqa} with intuitive physics~\cite{battaglia2016interaction} and the future
state anticipation tasks~\cite{finn2016unsupervised}.

\subsection{Table-top Interaction Visual \emph{What-If} Questions~(TIWIQ) Dataset}
\label{sec:tiwiq}

Existing Visual QA datasets~\cite{malinowski2014multi,ren2015imageqa,antol2015vqa} focus on static scenes, whereas datasets commonly used in future prediction tasks such as CityScapes~\cite{Cordts2016Cityscapes} involve a passive observer (future states are not conditioned on the agent's action). 
Since we are interested in the question answering task involving ``physical intelligence'', we collect
a new table-top interaction visual \emph{what-if} questions~(TIWIQ) dataset. This dataset has synthetic table-top scenes, with pairs of action descriptions and ground-truth descriptions of the outcomes of the specified action. We stick to synthetic scenes and a physics simulation engine to build this dataset as it provides physics, and enables controlled experimentation.

\noindent
\textbf{Scenes:}
To obtain the TIWIQ dataset we instantiate random table-top scenes in a physics engine, simulate actions on these scenes and collect human annotations describing the actions and the consequences. 
Each training sample in the TIWIQ dataset contains a table-top scene with five objects, each randomly placed upon the table. 
The five objects are chosen from eight items from the YCB Object Dataset~\cite{ycbdataset}: a foam brick, a cheez-it box, a chocolate pudding box, a mustard bottle, a banana, a softball, a ground coffee can, and a screwdriver.

\noindent
\textbf{Actions:}
A random action is chosen to be performed on a single random object, simulated using the Bullet 3~\cite{bullet3} physics engine. The resulting trajectories are rendered into a video of the interactions. The actions can be one of four:
\begin{enumerate*}
	\item Push an object in a specific direction.
    \item Rotate an object clockwise or anti-clockwise.
    \item Remove an object from the scene.
    \item Drop an object on another object.
\end{enumerate*}

\noindent
\textbf{Annotation:}
The objects shown in rendered videos have colored outlines, and when questions are posed to annotators, objects are referred to by their outline color rather than their name. This avoids the questions biasing the annotator's vocabulary with regard to object names.

\noindent
\textbf{Human Baseline:}
We have also collected a human performance benchmark on the visual what-if question answering task on the TIWIQ dataset. To obtain the human performance baseline, the annotators were shown a static image of the scene and a description of the action to be performed and were asked to describe what happens to different objects in the scene. We compare the performance of the model proposed in section~\ref{sec:proposedModel} to this human performance benchmark.

\noindent
\textbf{Dataset Statistics:}
We have generated and annotated 15 batches of data. Each batch has 17 examples of each action, totaling 68 examples per batch. In total, we have 1020 annotated examples. Three batches, totaling 204 examples and 20\% of the dataset, are dedicated to testing. For each scene, there are four generated descriptions (one for each object that is not being acted on), therefore there are 4080 ($1020 * 4$) annotated descriptions. However, descriptions relate to movement or interactions between objects only around 25\% of the time. This is due to the random placement of objects sometimes resulting in scenes with spatially separated objects, and therefore some actions having no impact on most objects in a scene. This results in approximately 1000 movement and interaction descriptions across the dataset. Only these annotations are used to train the description generation model.

\noindent
\textbf{Vocabulary Statistics:}
The vocabulary of the dataset is explored by counting the number of unique words used across the dataset (1-gram), as well as the number of unique n-grams for values 2 to 5 (2,3,4,5-grams). This is shown in Table~\ref{table:data_stats}. These statistics are reported for the action description annotations, the action effect annotations, and the two together. It is worth noting that the vocabulary of the action description dataset is significantly smaller than the vocabulary of the effect description dataset. This is due to the range of actions being specified by the design of the scenario, while the range of effects has no such constraints.

\begin{table}[h]
\begin{center}
\caption{The size of the vocabulary for the action and effect descriptions and the whole TIWIQ dataset, including the average sentence length and the number of unique n-grams in each subset of the dataset.}\label{table:data_stats}
\begin{tabular}{lllllll}
\hline\noalign{\smallskip}
  Descriptions & Length & 1-gram & 2-grams & 3-grams & 4-grams & 5-grams \\
\noalign{\smallskip}
\hline
\noalign{\smallskip}
  Action & 9.63  & 107 & 323 & 565   & 757   & 867 \\
  Effect & 7.663 & 110 & 403 & 724   & 981   & 1,075 \\
  All     & 8.582 & 152 & 619 & 1,171 & 1,653 & 1,895 \\
\hline
\end{tabular}
\end{center}
\end{table}

\section{Our Model}
\label{sec:proposedModel}
Recent advances in Deep Learning architectures have dominated Visual QA and image captioning tasks. The dominant approach is to use end-to-end trainable neural networks which take the inputs, e.g. image and question, and predict the answer. The whole model is learned purely from the data. Driven by sizable datasets, they have outperformed previous purely model-based approaches e.g. relying on semantic parsing or rule-based systems \cite{malinowski2014multi,malinowski2017ask}. Although latest work has also shown early success at applying this end-to-end machine learning paradigm to predicting future states of a scene, the time horizon remains limited to a few frames and rich object interactions are typically not considered, or the scenes are greatly simplified \cite{bhattacharyya2016long,watters2017visual,li2017visual}. Therefore, we argue for a hybrid model. We use a physics engine as a backbone in order to simulate outcomes of hypothetical actions, but we embed the simulation into learning-based components that drive the simulation as well as interpret its outcome in terms of a natural language output.

\subsection{Model Overview}
\begin{figure}[b]
\centering
\includegraphics[width=\textwidth]{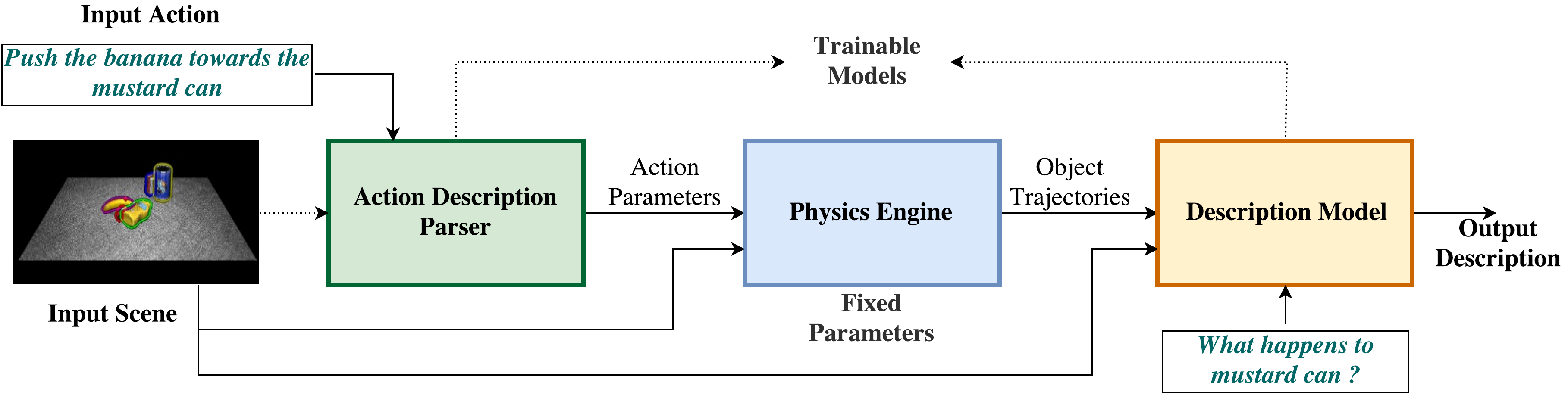}
\caption{Overall architecture of the proposed hybrid QA model.}
\label{fig:overallModel}
\end{figure}
The proposed hybrid question answering model consists of three distinct components as shown in figure~\ref{fig:overallModel}.
There are two inputs to the whole model.
The first  is a list of object types and their initial pose (position in 3-dimensional space, and a 3x3 rotation matrix) in the scene. We always assume the same table position for every case.
The second input is the action description. This was provided by human annotators, and describes some action performed on one of the objects in the scene, for example ``The robot pushes the mustard container to the left''.

Both inputs are used by a ``parser'' (we use a neural network as the parser) to extract parameters of the action to be performed. This includes parsing the action type, object to be acted upon and parameters of the action. This extracted information serves as an input to a physics engine, which then simulates the parsed actions on the input scene to produce trajectories for each object in the scene. While these trajectories encode everything that happened in the simulation, they are not human readable. The description model takes these trajectories as input and produces a natural language summary of the state of each object under the influence of this action. The action parser model and description models are comprised of neural networks and their parameters are learned from the data. The physics engine is model driven and has no trainable parameters.
In the following subsections we  discuss each of these components and how they interact in more details.

As well as being described in this document, all models are illustrated accurately in the corresponding figures. Each component in the illustrations describes a single layer, whether it is an RNN or a fully connected layer. All details of the layers are given in the supplementary material. This includes layer sizes, dropout, and activation functions.

\subsection{Action Description Parser}
\label{sec:extractingInformationModels}

\begin{wrapfigure}[26]{L}{0.45\textwidth}
\centering
\includegraphics[width=0.45\textwidth]{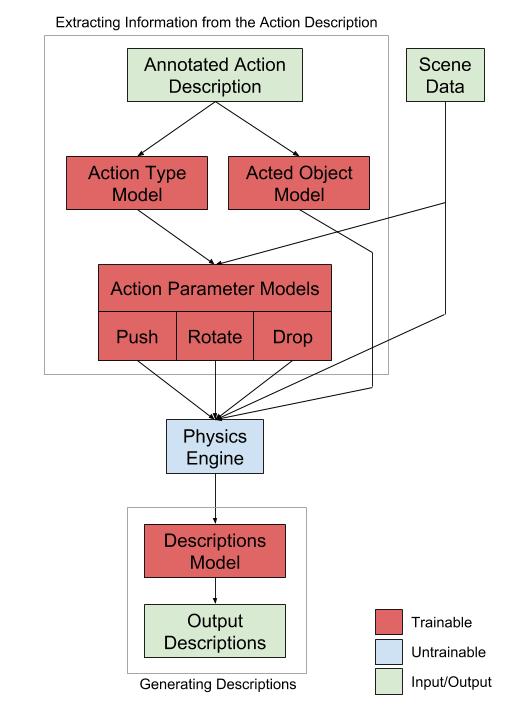}
\vspace*{-15pt}
\caption{Illustration of the interaction of subcomponents of the action description parser when inferring parameters of the action to be performed from input sentence.} \label{fig:end_to_end_model}
\end{wrapfigure}

The first step in the pipeline is parsing the exact nature of the action to be performed from the input sentence description. This model forms part of figure~\ref{fig:end_to_end_model}. It again consists of three components. First component is the \emph{Action Type Model} which infers the type of the action described in the input (push, rotate, drop or remove). This is a recurrent neural network (RNN) that embeds the tokenized action description, iterates over it using a long short-term memory (LSTM) model, and puts the final output of the LSTM through a fully connected layer with softmax activation. These outputs are treated as the probability that each action type was described in the action description.

The second component is the \emph{Acted Object Model}, which predicts the object in the input scene on which the described action is to be performed. The structure is identical to the Action Type Model, with the exception that it outputs probabilities of each class being the object to act upon.

Finally, the third component is a set of \emph{Action Parameter Models}, which infer the exact parameters of the action depending on the action type. Depending on the inferred action type, one of four things happens. If the action type is a push, rotate, or drop action, then the corresponding parameter model is called with the action description and the input scene. If it is a remove action, no parameters need to be inferred as the object is simply removed from the scene.

There are parameter models for three of the four actions: push, rotate, and drop. Each of these models use recurrent networks for embedding the action description specific to the action type.

The \emph{Push Parameter Model} infers the direction of the push by outputting a $(x,y)$ push direction vector in its final layer. The activation for this layer is sigmoid in order to cap both components from -1 to 1. When the physics engine simulates this push direction, the $(x,y)$ components are converted into an angle, removing the magnitude of the push.

The \emph{Rotate Parameter Model} is a binary classifier which predicts whether the rotation is clockwise or anti-clockwise, using softmax activation for classification.

The \emph{Drop Parameter Model} outputs a classification of which other object the acted object is dropped on. This also has a softmax activation for the classification, running over all possible objects.

Each component model that has text as input requires the text to be embedded. When it provided an improvement in performance, GloVe pre-trained word embeddings\cite{pennington2014glove} were used. Details on which layers used pre-trained embeddings, including the size of the embeddings, is given in the supplementary material.

\begin{wrapfigure}[27]{R}{0.45\textwidth}
\centering
\includegraphics[width=0.45\textwidth]{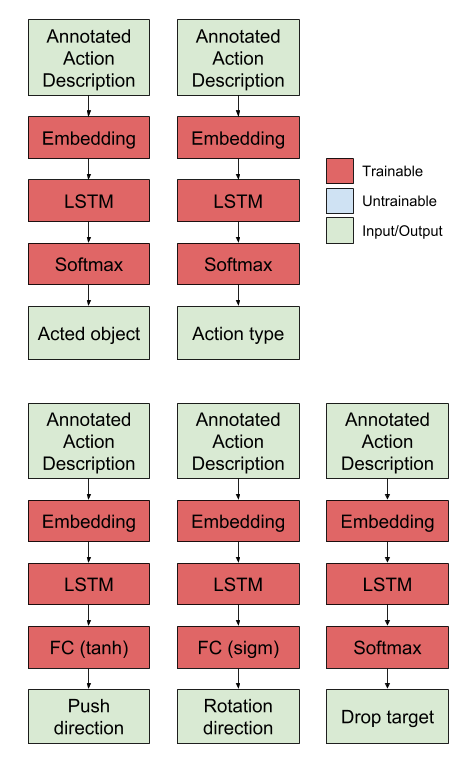}
\vspace*{-20pt}
\caption{Illustrations of the action parsing models. Between them, only the final fully connected (FC) layers differ.}\label{fig:parameter_models}
\end{wrapfigure}

\subsection{Physics Simulation}
The Acted Object Model extracts the action type, the object of interest, and the parameters of the action.
We use Bullet 3~\cite{bullet3} as a physics engine, with object representations from the YCB Object Dataset~\cite{ycbdataset}. We use object convex decompositions for collision detection which are calculated using VHACD, an algorithm included in the Bullet 3 source code. Pushes and rotations are implemented via impulses.

The physics engine is initialized with initial object poses. The engine is run for one second of simulation time in order for objects to settle in.

The inferred action is then performed on the inferred object, and the simulation is run for a total of five seconds at a sampling rate of 300Hz. Trajectories for each object are extracted from the simulation as a list of translation and rotation pairs, where the translation is a point in 3-dimensional space and the rotation is represented by a 3x3 rotation matrix.

We then run a simple algorithm to check if an object was affected by the action. To do this, we look at the trajectory for a single object, normalize the pose using a standard deviation and mean estimated from the entire training data set, and then calculate the standard deviation of both the translation and rotation, resulting in two floating point values. We say that an object was affected by the action if either of these values exceed a certain threshold. These thresholds were calculated by running a grid search over possible threshold values for either value, and picking the pair that resulted in the best classification accuracy on the training set.

\subsection{Generating Descriptions}
\begin{wrapfigure}[29]{R}{0.35\textwidth}
\centering
\includegraphics[width=0.35\textwidth]{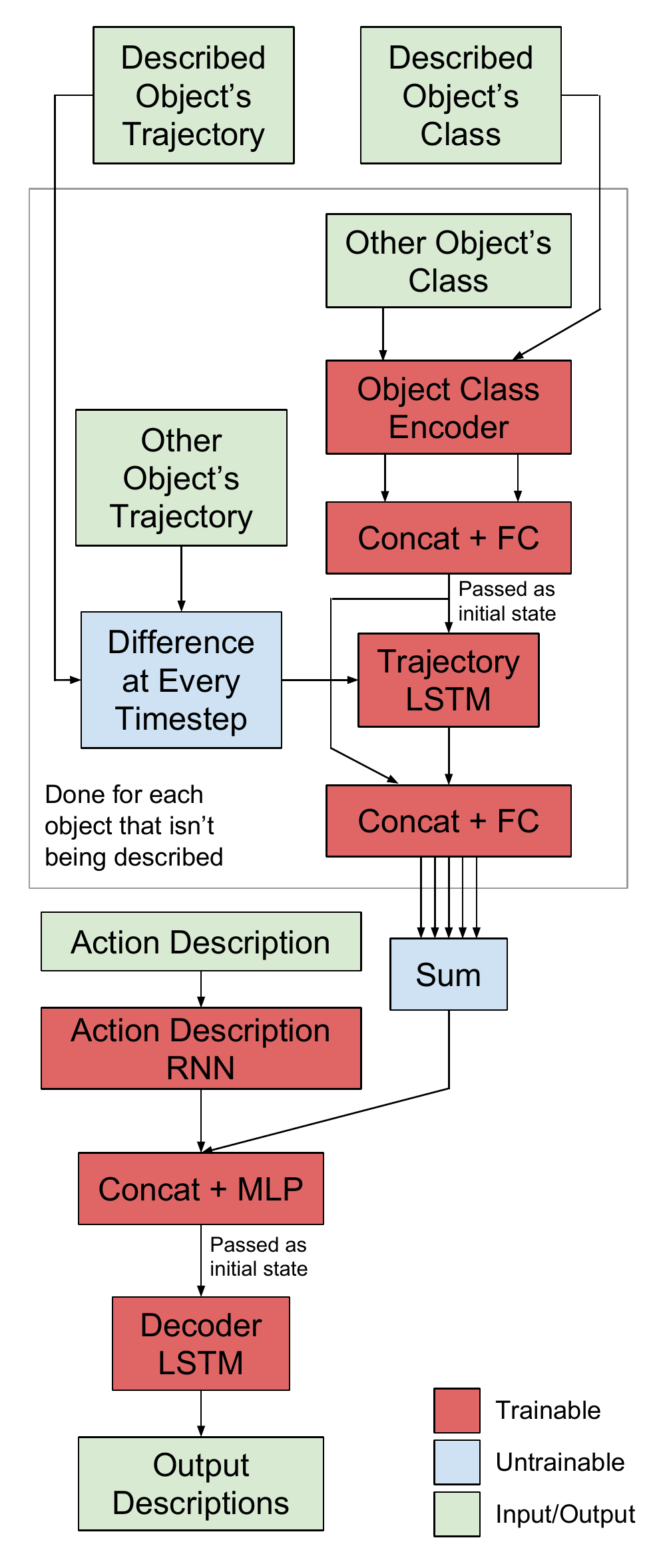}
\caption{An illustration of the description model. The section outlined in a gray square is run for every object that isn't the object that is currently being described. }
\label{fig:description_model}
\end{wrapfigure}

The \emph{Description Model} shown in figure~\ref{fig:description_model} uses the trajectories from the physics simulation to produce a one sentence summary of the effect of the action on each object in the scene. This model is run independently for each generated description with the following inputs:
\begin{enumerate*}
	\item The action description.
	\item The object class to describe.
    \item The trajectory of the object to describe.
    \item A list of other object classes in the scene.
    \item A list of other object trajectories in the scene.
\end{enumerate*}

The object classes are encoded as a one-hot vector, and the trajectories are encoded as a list of points in 3-dimensional space.

At training time, ground truth trajectories are used, but when the complete hybrid model is being evaluated, predicted trajectories are generated via physics simulation.

The description model works in two stages. First the input trajectories of the target object~(whose state is being described) and other objects in the scene are compared and trajectory embeddings are obtained. Then these trajectory embeddings and action description embeddings are input to a decoder LSTM language model, which generates the final description.

To obtain the trajectory embeddings we iterate over each of the other objects in the scene --- that is, the ones that are not currently being described. For each object, we compute the difference between its trajectory and the trajectory of the object to be described, at each time step. These difference vectors are then embedded and iterated over using an LSTM. The initial state of this LSTM is provided by embedding both of the object classes and putting them through a fully connected layer. The final hidden state of this LSTM should encode the interactions between the two objects. This output trajectory embedding is concatenated with the object encodings of the two relevant objects. We find that including these embeddings after as well as before the trajectory encoding LSTM improves the overall model's performance. 

The input action description is encoded using an LSTM (as in earlier models, such as the action description model). A fully connected layer is used to transform the concatenated trajectory embedding vector and the encoded input instruction into the right dimensions and is used to initialize the hidden state of the decoder LSTM.\@ The input for the decoder LSTM at time $t_0$ is the \emph{start of sentence} token, and the input at time $t_i$ is the output from $t_{i - 1}$. At each step the decoder LSTM outputs the next word and this repeats until the \emph{end of sentence} token is predicted. This process is carried out to generate a description for each object in the scene.

\subsection{Implementation Details}
We have implemented the hybrid model and all components in Python using the Keras~\cite{chollet2015keras} library with the TensorFlow~\cite{tensorflow2015-whitepaper} backend. For the description model, custom layers were introduced into Keras using TensorFlow. Overall runtime of the system is 1.76s, where prediction time of the Action Description Parser and Description Generation is negligible. Almost all time is spent in the simulation part. For reproducibility and to stimulate more work on this challenge, we will release code, models and setup.

\section{Results}
We evaluate our overall hybrid approach as well as the individual components on the proposed dataset as well as compare to an end-to-end learning and human baseline. We provide example results and analyses that highlight the benefits of our approach.

\subsection{Performance of Hybrid Model Components}

\begin{wraptable}[13]{R}{0.5\textwidth}
\centering
\vspace*{-15pt}
\caption{Comparing classification accuracy of neural network based and SVM based models on different tasks.\medskip}
\label{table:component_models}
\begin{tabular}{lll}
\hline\noalign{\smallskip}
	Task & NN & SVM \\
\noalign{\smallskip}
\hline
\noalign{\smallskip}
Action Type 			& \textbf{97.5\%} & 97\% \\
Acted Object  		    & \textbf{94\%} & 90\% \\
Push Parameters 		& \textbf{90\%} & 44\% \\
Rotation Parameters 	& 68\% & \textbf{72\%} \\
Drop Parameters 		& \textbf{90\%} & 36\% \\
\hline
\end{tabular}
\end{wraptable}

We separately evaluate the performance of the six components  of the hybrid model, using ground truth annotations at these intermediate stages. First, we show the performance of the action description information extraction models (action type model, acted object model, push / rotate / drop parameter model) in table~\ref{table:component_models}.  We created simple support vector machine~(SVM) baselines in order to benchmark the more powerful neural models. The input to these SVMs is a vector of word counts for every word in the data vocabulary. We find that in all cases bar one, the neural models significantly outperform  SVMs, as shown in Table \ref{table:component_models}. The exception is the rotation parameter model, which is outperformed by 5.8\%. The performance for the rotation parameter model is particularly poor due to noisy annotations in the cases of rotation actions. Through looking at a small subset of the rotation action annotations, we have found that 30-40\% of the annotations are mislabeled in some way --- either giving the wrong rotational direction, or annotated as a push action.

To compare the push parameter model with a classification network, we discretize the angle inferred by the neural model into eight directions (e.g. left, top-left, up). The SVM also classifies to one of those eight directions, allowing us to compare the performance of these two models.

\subsection{Quantifying the Hybrid Model Performance}
We will now quantify the performance of the proposed hybrid model and  baselines on the test set. 

\noindent\textbf{Metrics:} To measure the description performance we use the standard metrics used in evaluating image captioning models such as BLEU, CIDEr, ROUGE and a custom metric COM~(``Correct Objects Mentioned'') that we designed for this specific problem. This metric searches the descriptions for different object names, creating a list of objects mentioned in the text. This is done for both the prediction and the ground truth. The COM metric is computed as the intersection over union of these two sets. The upper-bound for COM is 1, and occurs when all correct objects are mentioned in all predictions for a scene. Image captioning metrics such as BLEU focus on overall n-gram matching of the generated description with the ground truth, regardless of the importance of each word, whereas COM directly measures how well models identify the acting objects in a scene.

\begin{figure}
\centering
\includegraphics[height=4cm]{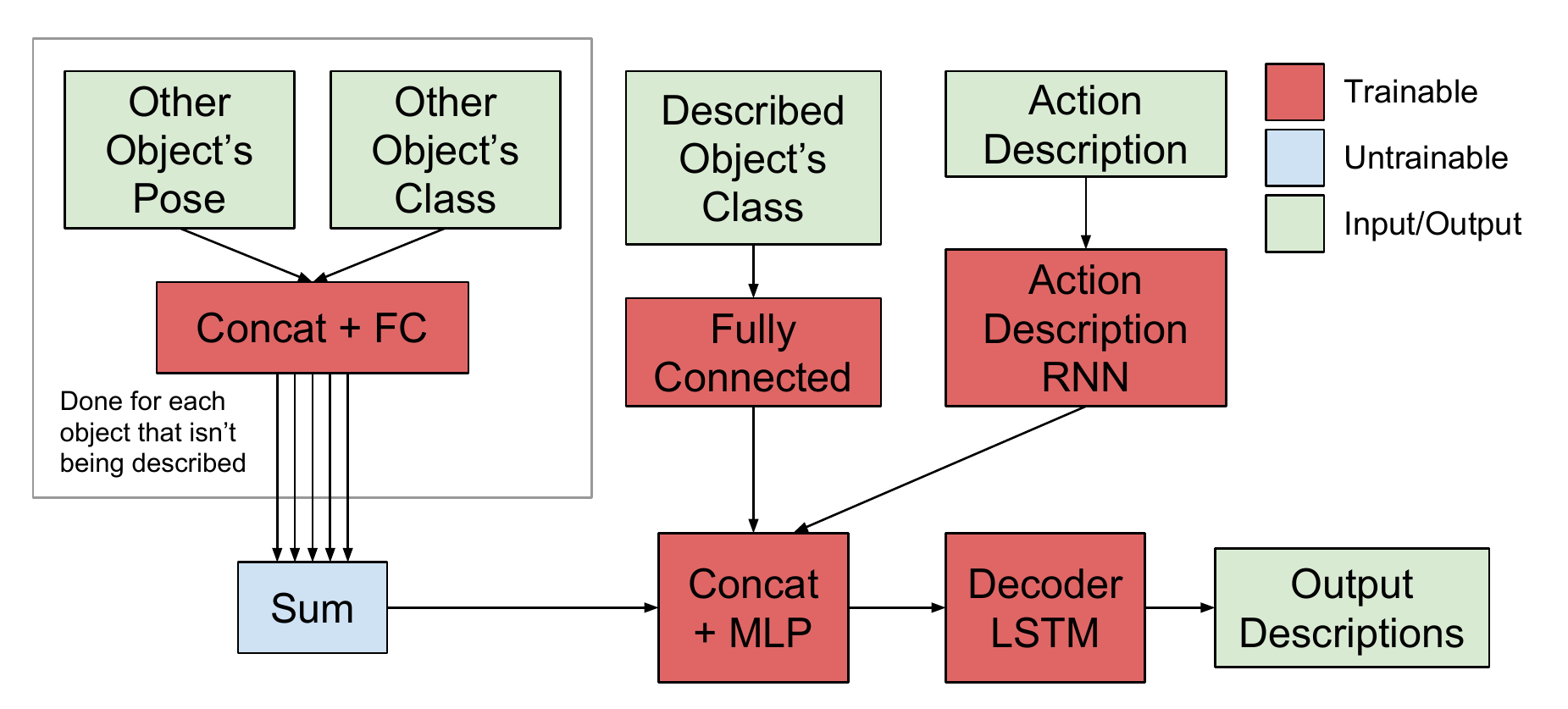}
\caption{Illustration of the data-driven model.}
\label{fig:data_driven_model}
\end{figure}

\noindent\textbf{Hybrid Model Compared to Baselines:}
We compare the performance of the hybrid model against three other baselines on the test set. The first is a pure data-driven model, an end-to-end trainable neural network illustrated in figure \ref{fig:data_driven_model}. The inputs to this network are the input action description, the initial scene, and the object to describe. The action description is embedded and then run through an LSTM and the final output of this LSTM is taken. Each class and pose in the initial scene is flattened into a vector, and each of these is put through a fully connected layer and summed together. The object to describe is encoded as a one-hot vector and passed through a fully connected layer. Each of these encodings are concatenated together and treated as input for the decoder LSTM, which generates a description for the specified object class.

\noindent\textbf{Human Baseline and ``Upper Bound'':}
The second set of descriptions is from a human baseline, mentioned in section \ref{sec:tiwiq}. Human annotators were shown the input scene and action description, but not the video of the action taking place. They were asked to describe what happens to each object. This simulates the same task tackled by the hybrid, and pure data-driven models.
Finally, the third baseline is obtained by feeding the ground-truth trajectories to the description model. This represents an upper-bound on the hybrid model performance.

\noindent\textbf{Discussion:}
We find that the hybrid model outperforms the data-driven model in all metrics, with an increase of 15.4\% in the BLEU metric, and an increase of 20.5\% in the COM metric, as illustrated in table \ref{table:descGenModels}. This provides evidence for incorporating a physics engine for solving physics based anticipation problems over a pure data-driven approach. The performance of the hybrid model is close to its upper-bound description model and this gap comes from the cases where the action parsing model failed. However, there is still a gap in performance in terms of the COM metric between the proposed hybrid model and the human benchmark, indicating the scope for future improvements.

\noindent\textbf{Human Baseline Discussion:}
Under most metrics, the hybrid model outperforms the human baseline. However, this is misleading: the human baseline contains high-quality annotations, and under domain-specific metrics such as COM, is evaluated with a near-perfect score (0.953). Its large error in BLUE, CIDEr, and ROUGE results from the differing vocabularies between the human baseline and the ground truth. For this reason, comparison between the hybrid model and the human baseline is difficult to achieve using these metrics; a similar problem is common in the image captioning domain.

\begin{table}
\begin{center}
\caption{Comparison of description generating models. Best values between the hybrid model and data-driven model are highlighted. This shows that the hybrid model exceeds the data-driven model and even the human baseline in some metrics.}
\label{table:descGenModels}
\begin{tabular}{lrrrrrrrr}
\hline\noalign{\smallskip}
	Model & BLEU & BLEU-1 & BLEU-2 & BLEU-3 & BLEU-4 & CIDEr & ROUGE & COM \\
\noalign{\smallskip}
\hline
\noalign{\smallskip}
Description Model  & 0.280 & 0.439 & 0.345 & 0.203 & 0.133 & 1.118 & 0.421 & 0.671 \\
Human Baseline	    & 0.191 & 0.207 & 0.192 & 0.186 & 0.181 & 1.849 & 0.209 & 0.953 \\
\hline
Hybrid Model       & \textbf{0.262} & \textbf{0.407} & \textbf{0.322} & \textbf{0.184} & \textbf{0.134} & \textbf{1.118} & \textbf{0.396} & \textbf{0.640} \\
Data-Driven Model  & 0.227 & 0.375 & 0.282 & 0.154 & 0.099 & 0.896 & 0.376 & 0.531 \\
\hline
\end{tabular}
\end{center}
\end{table}

\subsection{Qualitative analysis}

We provide qualitative examples and analysis in table \ref{table:exampleScenes}. In these examples, the hybrid model can be seen generating more specific and accurate descriptions of the results of actions compared to the data-driven model. There are three main failure cases of the data-driven model.

The first of these is shown in row 1 of table \ref{table:exampleScenes}. In this example, the hybrid model correctly predicts the object which hits the foam (in this case the screwdriver) while the data-driven approach predicts that a different object in the scene will hit the foam. Accuracy is lost here due to the data-driven model not being able to reliably infer the object that interacted with the subject object. Our hybrid model performs better in this case, presumably because it was able to use the trajectories from the physics engine to infer the correct object.

The second main failure case is shown in row 2. Both models are correct but the hybrid model gives a more precise description, stating correctly which object hit the mustard container. The data-driven model gives a more vague description by not stating the acting object and just describing the movement.

The third failure case is shown in row 3. Often, the data-driven model produces a description where both the object being acted on and the object affecting it are the same. This could be due to the data-driven model making the best guess it can --- if it knows that the class ``screw driver'' appears in the text but does not know what the other object could be, and it knows that the sentence should reference two objects, then it may choose to mention ``screw driver'' twice in the sentence. This failure case, although more prevalent in the data-driven model, shows up in the hybrid model too as seen in row 4 of table \ref{table:exampleScenes}.

There is a failure case unique to the hybrid model. The data-driven model was trained only on cases where the action did have an effect on the object. However, the hybrid model has to infer whether there was an effect. This results in some cases where the hybrid model misclassifies the object as ``not moving'' and generates the ``nothing'' description. An example of this case is shown in row 5 of table \ref{table:exampleScenes}.

\subsection{Ablation Analysis}
We also analyze the error introduced by the different components within the hybrid model. We do this by introducing, one-by-one, the ground truth values for a particular component instead of the predicted values. The results of this are shown in table \ref{table:withTruths}. We can see that introducing the ground-truth for whether an object moved provides the biggest increase in performance, implying that the hybrid model loses a lot of accuracy when predicting whether an object moved. Conversely, we can also see that the Action Type and Acted Object models introduce relatively small amounts of error, suggesting they correctly model the ground truth.

\begin{table}[t]
\begin{center}
\caption{Comparison of how the performance of the hybrid model improves when cumulatively adding truth values for each of the components.}
\label{table:withTruths}
\begin{tabular}{lllll}
\hline\noalign{\smallskip}
	Model & BLEU & CIDEr & ROUGE & COM \\
\noalign{\smallskip}
\hline
\noalign{\smallskip}
All Predictions:               & 0.262 & 1.118 & 0.396 & 0.640 \\
With True Action Type:         & 0.264 & 1.126 & 0.398 & 0.644 \\
...and True Acted Object:      & 0.265 & 1.129 & 0.400 & 0.646 \\
...and True Action Parameters: & 0.272 & 1.153 & 0.406 & 0.662 \\
...and True Trajectories:      & 0.268 & 1.086 & 0.401 & 0.645 \\
...and True Object Acted On:      & 0.283 & 1.133 & 0.424 & 0.673 \\
\hline
\end{tabular}
\end{center}
\end{table}

\section{Conclusion}
We have proposed a new task that combines scene understanding with anticipation of future scene states. We argue that this type of ``physical intelligence'' is a key competence of an agent that is interacting with an environment and tries to evaluate different alternatives. In contrast to prior work on quantitative predictions of future states, here, we focus on a qualitative prediction that describes the outcome for a certain object with a natural language utterance. 
Owing to such a formulation, we can train and evaluate our agent on long-term future anticipation, where the model can easily ignore irrelevant details of the scene or the interactions. This contrasts with future frame synthesis where all the details have to be correctly modeled.

Due to the lack of suitable datasets, we introduced the first dataset and an evaluation protocol for this challenging task. Our proposed model is the first that combines a question answering architecture with a physics engine and verbalizes different outcomes dependent on the visual and language input. In particular, our hybrid model outperforms a purely data-driven  Deep Learning baseline. We believe that such hybrid models that combine a complex simulation engine with data-driven components represent an exciting avenue for further research as they allow for generalization, scalability and improved performance in challenging scenarios with a high combinatorial complexity. 

\begin{table}
\centering
\caption{Examples of different scenes in the data set with annotations from ground truth, human baseline, the predictions from the hybrid model, and predictions from the baseline data-driven model. The hybrid can be seen giving more precise descriptions than the data-driven model.}
\label{table:exampleScenes}
\resizebox{10cm}{!}{
\begin{tabular}{l|l}
\hline\noalign{\smallskip}
	Input Scene And Action & Output Description \\
\noalign{\smallskip}
\hline
\noalign{\smallskip}
\begin{tabular}{p{5cm}}
  \includegraphics[width=5cm, trim={0.5cm 2cm 0 2cm}, clip]{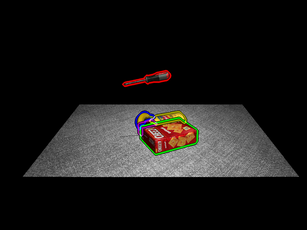}
  1) ``the robot drops the screw driver on the foam'' --- what happens to the foam?
\end{tabular} &
\begin{tabular}{p{7cm}}
  \textbf{Ground Truth} ``the foam is pushed a little by the screw driver'' \\
  \textbf{Human Baseline} ``the foam is pushed because the  screw driver drops on it'' \\
  \textbf{Hybrid Model Prediction} ``the foam is pushed by the screw driver'' \\
  \textbf{Data-Driven Model Prediction} ``the foam is pushed by the mustard container'' \\
\end{tabular} \\
\hline
\begin{tabular}{p{5cm}}
  \includegraphics[width=5cm, trim={0.5cm 2cm 0 2cm}, clip]{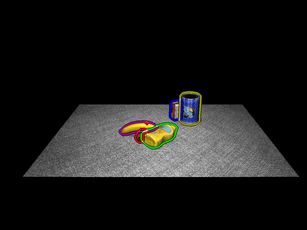}
  2) ``the robot spins the screw driver in anti-clowise direction'' --- what happens to the mustard container?
\end{tabular} &
\begin{tabular}{p{7cm}}
  \textbf{Ground Truth} ``the screw driver pushes the mustard container'' \\
  \textbf{Human Baseline} ``the mustard conatainer moves a little due to the impact of spinning screw driver'' \\
  \textbf{Hybrid Model Prediction} ``the screw driver pushes the mustard container'' \\
  \textbf{Data-Driven Prediction} ``the mustard container shakes a little from the impact'' \\
\end{tabular} \\
\hline
\begin{tabular}{p{5cm}}
  \includegraphics[width=5cm, trim={0.5cm 2cm 0 0cm}, clip]{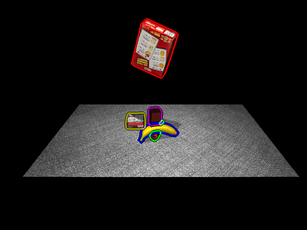}
  3) ``the robot drops the cheese box on the foam'' --- what happens to the screw driver?
\end{tabular} &
\begin{tabular}{p{7cm}}
  \textbf{Ground Truth} ``the screw driver is pushed by the cheese box'' \\
  \textbf{Human Baseline} ``the screw driver is pushed by the cheese box'' \\
  \textbf{Hybrid Model Prediction} ``the screw driver is pushed by the cheese box'' \\
  \textbf{Data-Driven Prediction} ``the screw driver is pushed by the screw driver'' \\
\end{tabular} \\
\hline
\begin{tabular}{p{5cm}}
  \includegraphics[width=5cm, trim={0.5cm 2cm 0 3cm}, clip]{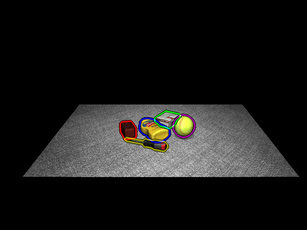}
  4) ``the robot spins the screw driver in anti-clockwise direction'' --- what happens to the chocolate box?
\end{tabular} &
\begin{tabular}{p{7cm}}
  \textbf{Ground Truth} ``the chocolate box is pushed by the screw driver'' \\
  \textbf{Human Baseline} ``the chocolate box is pushed a slightly by the screw driver'' \\
  \textbf{Hybrid Model Prediction} ``the chocolate box pushes the chocolate box'' \\
  \textbf{Data-Driven Prediction} ``the chocolate box is pushed by the screw driver'' \\
\end{tabular} \\
\hline
\begin{tabular}{p{5cm}}
  \includegraphics[width=5cm, trim={0.5cm 2cm 0 3cm}, clip]{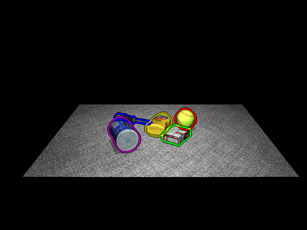}
  5) ``the robot  pushes the baseball to the middle of the table'' --- what happens to the chocolate box?
\end{tabular} &
\begin{tabular}{p{7cm}}
  \textbf{Ground Truth} ``the chocolate box is pushed by the baseball'' \\
  \textbf{Human Baseline} ``the chocolate box is pushed by the baseball'' \\
  \textbf{Hybrid Model Prediction} ``nothing'' \\
  \textbf{Data-Driven Prediction} ``the chocolate box is pushed by the chocolate box'' \\
\end{tabular} \\
\hline
\begin{tabular}{p{5cm}}
  \includegraphics[width=5cm, trim={0.5cm 8cm 0 13cm}, clip]{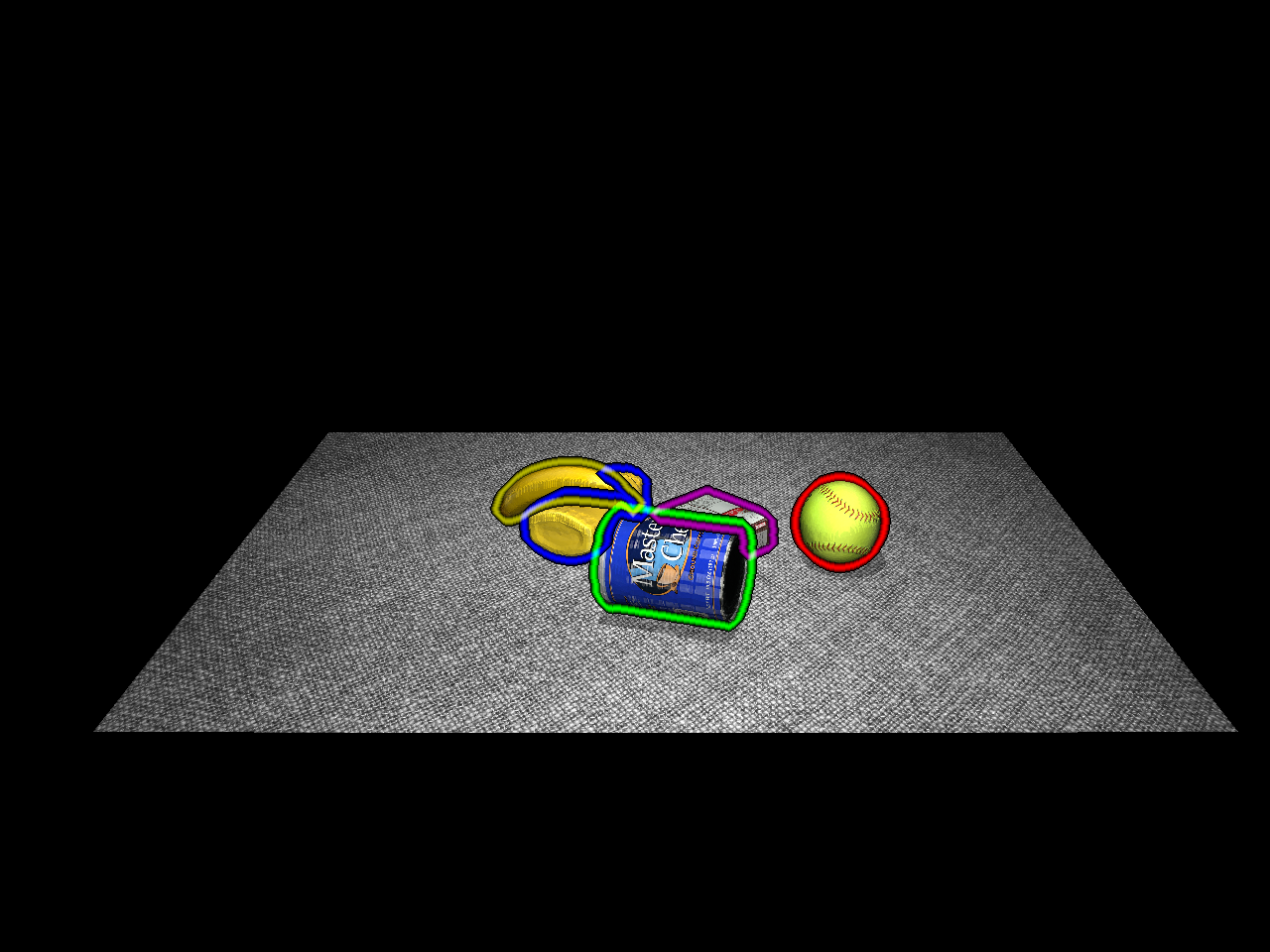}
  6) ``the robot rolls the baseball to the north-west side of the table and it drops off'' --- what happens to the banana?
\end{tabular} &
\begin{tabular}{p{5cm}}
  \textbf{Ground Truth} ``nothing'' \\
  \textbf{Human Baseline} ``nothing'' \\
  \textbf{Hybrid Model Prediction} ``nothing'' \\
  \textbf{Data-Driven Prediction} N/A \\
\end{tabular} \\
\hline
\end{tabular}
}
\end{table}

\section*{Acknowledgements}
We acknowledge MoD/Dstl and EPSRC for providing the grant to support the UK academics’ involvement in a Department of Defense funded MURI project through EPSRC grant EP/N019415/1.

\bibliographystyle{splncs}
\bibliography{ms}
\clearpage
\end{document}


\pagestyle{headings}
\mainmatter
\def\ECCV18SubNumber{}  

\title{Answering Visual {\em What-If} Questions:\\From Actions to Predicted Scene Descriptions\\Supplementary Material}

\titlerunning{Answering Visual {\em What-If} Questions}

\authorrunning{M. Wagner et al.}
\author{Misha Wagner\inst{1}\thanks{These authors contributed equally to this work.} \and Hector Basevi\inst{1}\printfnsymbol{1} \and 	Rakshith Shetty\inst{2} \and Wenbin Li\inst{2} \and \newline Mateusz Malinowski\inst{2} \and Mario Fritz\inst{3} \and Ales Leonardis\inst{1}}
\institute{University of Birmingham \and Max Planck Institute for Informatics, Saarland Informatics Campus \and CISPA Helmholtz Center i.G., Saarland Informatics Campus}

\maketitle

\begin{subappendices}
\renewcommand{\thesection}{\Alph{section}}

\section{Data set and model output variation}

We evaluate the vocabulary of the dataset and the vocabulary of the model output by forming ``multi-level pie charts'' to display the distribution of vocabulary used, showing only symbols taking up over 0.1\% of the pie chart. These are shown in Figure~\ref{fig:vocabulary_comparison} of the supplementary material.

The visualisation shows a variety in sentence structure for the data set. However, the model learns the most common sentence structures in order to describe the effect of an action on an object, losing some of the variation in the human vocabulary.

\begin{figure}[ht]
\centering
\includegraphics[width=\textwidth]{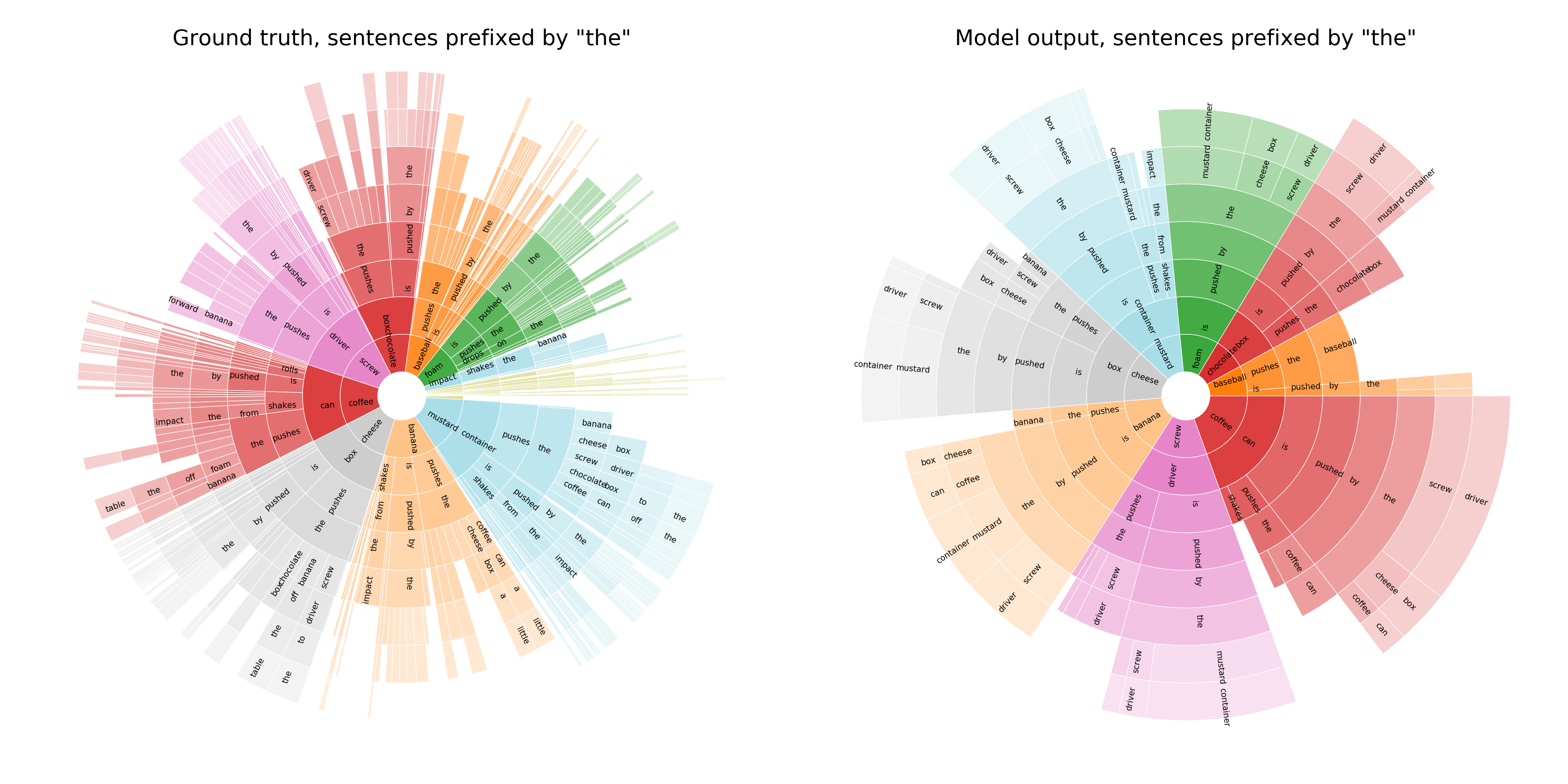}
\caption{Comparison between the vocabulary of the ground truth effect descriptions and the hybrid model output. Segment circumference represents frequency of the word occurring after the parent segment in the inner layer. Instances of ``the'' as the first word in a sentence have been removed (a negligible amount of descriptions do not start with ``the''), and visualisation depth is capped at eight layers/words.}
\label{fig:vocabulary_comparison}
\end{figure}

\section{Replication details for networks}

The majority of important implementation details are shown in Figure~2, Figure~3, and Figure~4 of the paper. They describe the flow of information inside and between each component of the hybrid model but abstract away key features, such as model hyperparameters and activation functions. These are shown in Table~\ref{table:hyperparameters} of the supplementary material.

For loss functions, all classification models (action type, acted object, push rotate parameters, and drop parameters) use categorical cross entropy, while other models (push parameters and description model) use mean squared error. All models were terminated after 10 consecutive epochs of no improvement in validation error.

\begin{table}[ht]
\begin{center}
\caption{Each component model's hyperparameters, including dimension size, activation function, and level of dropout.}\label{table:hyperparameters}
\begin{multicols}{2}
  \begin{itemize}
    \item Action type model
      \begin{itemize}
        \item Use pre-trained word embeddings
        \item Word embedding size of 200
        \item LSTM size of 240
        \item No dropout
        \item Softmax activation for probabilities over action types
      \end{itemize}

    \item Drop parameters model
      \begin{itemize}
        \item Use pre-trained word embeddings
        \item Word embedding size of 100
        \item LSTM size of 55
        \item No dropout
        \item Softmax activation for probabilities over object types
      \end{itemize}

    \item Rotate parameters model
      \begin{itemize}
        \item Use pre-trained word embeddings
        \item Word embedding size of 200
        \item LSTM size of 119
        \item Dropout of 0.11
        \item \emph{sigmoid} activation for binary output (clockwise or anti-clockwise)
      \end{itemize}

    \item Push parameters model
      \begin{itemize}
        \item Do not use pre-trained word embeddings
        \item Word embedding size of 100
        \item LSTM size of 286
        \item No dropout
        \item \emph{tanh} to obtain $x, y \in [-1, 1]$ push direction
      \end{itemize}

    \item Description model
      \begin{itemize}
        \item Trajectory encoding of 512
        \item Trajectory LSTM size of 32
        \item No encoder dropout
        \item Decoder LSTM size of 68
        \item Decoder dropout of 0.03
        \item Object encoding size of 512
        \item 30 trajectory samples
        \item Action description encoding size of 512
        \item Action description LSTM size of 32
      \end{itemize}
  \end{itemize}
\end{multicols}
\end{center}
\end{table}

\section{Scalability of physics engines}
The proposed system incorporates a physics engine for physical prediction. The input to this engine is a set of object poses and a set of action parameters, and the output is a set of trajectories for the objects. Physics engines provide a number of advantages:
\begin{itemize}
\item Principled numerical approximation of Newtonian physics.
\item Natural input and output representations.
\item Generalisation to arbitrary object shapes and properties.
\item Zero learned parameters inside the engine.
\end{itemize}
However, collision detection within these physics engines can be an expensive step, particularly for high-detail shape representations and complex geometries. Consequently these engines can scale poorly with increasing scene complexity. However, there are a number of techniques that can be employed to improve computational efficiency:
\begin{itemize}
\item Complex geometries can be approximated by convex decompositions.
\item Simple geometries can be approximated by collections of 3D primitives.
\item The temporal resolution can be optimised to balance computational efficiency and accuracy.
\item Computation can be parallelised and executed on a multi-core CPU or GPU.
\end{itemize}
These techniques mean that the proposed system can be applied to and would scale to more complex scenarios and scenes without a prohibitive drop in performance.
\end{subappendices}